# TurKPF: TurKontrol as a Particle Filter

**Ethan Petuchowski**[1] and **Matthew Lease**[2]

[1]University of Texas at Austin, Department of Computer Science, ethanp@utexas.edu

[2]University of Texas at Austin, School of Information, ml@ischool.utexas.edu

**Abstract**

TurKontrol, an algorithm presented in (Dai *et al.* 2010), uses a POMDP to model and control an iterative workflow for crowdsourced work. Here, TurKontrol is re-implemented as "TurKPF," which uses a Particle Filter to reduce computation time & memory usage. Most importantly, in our experimental environment with default parameter settings, the action is chosen nearly instantaneously. Through a series of experiments we see that TurKPF and TurKontrol perform similarly.

## Introduction

Over the past several years, crowdsourcing has become a cheap way to hire people to process data. People are often hired to perform simple tasks that cannot be done with artificial intelligence, with many applications in both academia and commerce.

At this stage, a few questions discussed in the crowdsourcing research are: "What are the tasks whose costs can be effectively lowered with crowdsourcing?", "How can we ensure the data we get is reliable?", and "What specific methods/workflows of crowdsourcing can we use to do tasks well and reliably?"

As will be discussed in greater detail, TurKontrol (Dai et al. 2010) was presented as a way of bringing greater automation to the implementation and online-administration of crowdsourcing workflows via a tool from Artificial Intelligence called a Partially Observable Markov Decision Process (POMDP).

As is often the case in even POMDPs with small state-spaces (Pineau et al. 2006), Dai et al. note that "In a general model such as ours, maintaining a closed form representation for all these continuous functions may not be possible" (Dai et al. 2010, pg. 1171).

A *particle filter* approximates a real-valued distribution by storing samples from it (the "particles") rather than parameters of the underlying distribution. More particles in a given range of values represent a greater probability density in that range. In this way, it can be shown that as $N \to \infty$, the particle filter looks exactly like the original distribution (Donmez et al. 2010). To update the distribution after an action or an observation, a stochastic update function is applied to each particle, and the new set of particles represents the new belief-state. A benefit of the algorithm is that most computational effort is focused on the regions of the belief state with the highest density (those containing the most particles), and these are often also the regions we are most concerned with approximating accurately (Thrun et al. 2005).

We re-implemented TurKontrol as a particle filter called TurKPF[1] to understand and evaluate its performance. The main performance-characteristics of this implementation conform to those of the original version, and the cause of the one noted discrepancy remains unclear. TurKPF is able to compute the utility-maximizing action nearly instantaneously, and is highly tuneable to its environment. Although the current implementation is not tuned to any particular real-situation, the appropriate parameters can be found using machine learning techniques outlined in (Dai et al. 2013).

In the interest of brevity, we will refer to (Dai et al. 2010) with the abbreviation *DTC*.

## Related Work

A wide array of tasks have been successfully crowdsourced. Twitter uses crowdsourcing to teach its algorithms about the latest cultural trends (Chen 2013). By asking crowdworkers to do simple tasks, Snow et al. obtained data to train machine learning algorithms for Natural Language Processing at a similar quality and lower cost than using trained experts (Snow et al. 2008). Also, a cost-efficient method for doing SQL-style joins on big datasets of photos according to criteria such as "Do these two photos depict the same person?" was demonstrated in (Marcus et al. 2012). Crowdsourcing has been used experimentally to write captions for photos and even write entire articles (Kittur et al. 2011). It has also been used to research options for big decisions, such as evaluating which car to buy (Kittur et al. 2011).

---

[1] github.com/ethanp/crowdsourcing/tree/master/TurKPF

Ensuring reliability of the data is difficult because per-task compensation is often in the range of a few cents, and since tasks are not always simple or interesting, incoming data can be of low quality. One way to mitigate the effect of low quality data is to recruit redundant labelers to compensate for any individual worker's unreliability.

The simplest algorithm for resolving the answers from redundant labelers into a decision is to use the *majority vote*. Some improvements on this method focus on identifying who the reliable workers are, and diagnosing biases in individual workers toward particular mislabels. Techniques from statistics and machine learning have been applied to model workers' ability and figure out who's opinions to trust (e.g. Wang et al. 2011, Donmez et al. 2010). Many proposed algorithms for accomplishing this in an offline setting were implemented, compared, and contrasted in (Sheshadri and Lease 2013).

Certainly, there are many tasks which have already been successfully crowdsourced, but some predict that crowdsourcing can be applied to a much broader class of problems (Kittur et al. 2013). Creating *job-flows* has been proposed as a way of quickly and effectively tackling problems that one person cannot solve. One part of a job-flow might involve *iterative improvement*. This means asking one worker to start the task, then asking another worker to improve upon the last worker's response, and repeating until the task is sufficiently complete. One advantage to this workflow is that it does not rely on one worker's capability and patience to complete the entire task on his or her own. One disadvantage is that one must wait for the first worker to submit his or her work before enlisting the help of the second.

First we introduce POMDPs and TURKONTROL. Then we describe how TURKPF modifies TURKONTROL to be a particle filter. After that, we describe aspects of the performance of TURKPF. Finally, we conclude and offer suggestions for future work.

## Review of TURKONTROL, a POMDP

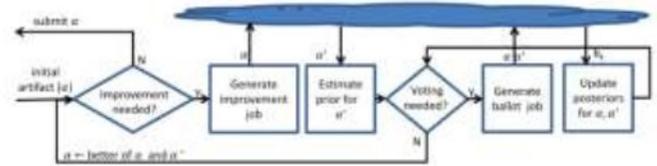

**Figure 1:** The mind of TURKONTROL; the blue blob is "the cloud." Reprinted from DTC.

## Motivation

Map-Reduce is a pattern for structuring programs that has been successfully applied to web-indexing and data analysis to facilitate fault-tolerant parallel processing of data on computer clusters (Dean and Ghemawat 2008). In Crowd-Forge, the authors proposed using concepts from Map-Reduce to parallelize crowdsourcing (Kittur et al. 2011). Specifically, they split (*map*) tasks up into pieces that do not rely on each other and post all those to the Web at the same time to be completed simultaneously. Then they use another task to aggregate (*reduce*) the results of those tasks. Afterwards, they may repeat this process as necessary with patterns of arbitrary complexity.

To automate the administration of such complex workflows, TURKIT (Little et al. 2010) is a job-flow designer that uses the Amazon Mechanical Turk (AMT) application programming interface (API) to incorporate crowdsourced data into a running program. However, there are certain constants one must specify during the initialization of this program that may lead to inefficient task allocation. For example, one must set the number of times to collect votes between iterations, the number of times to iterate before settling on an end-product, and there is no way of estimating the accuracy of the ballots you receive (Dai et al. 2010).

Working off of the TURKIT concept, TURKONTROL uses a Partially Observable Markov Decision Process (POMDP) to maintain an estimate of the current quality of the item being developed by the crowdsourced workers in an attempt to decide *dynamically/online* the parameters that TURKIT requires to be hard-wired. Specifically, when to assign a task to improve the item, when to collect votes on whether the improved item is better than the older version as well as how many votes to collect for each version, and when the item is of high enough quality to submit.

## Overview of POMDPs

TURKONTROL models iterative crowdsourcing tasks as a Partially Observable Markov Decision Process (POMDP).

This is a Markov model for situations in which the variables that one wants to estimate/maximize are not directly observable, but other observable variables allow the use of a Bayes' filtering algorithm to derive a *belief state*, i.e. likelihood estimates of each possible underlying world state.

The goal of an *agent* in a POMDP is to *maximize utility* given its *environment*. To set up a POMDP, we start by modeling the environment with sets of *states* $S$, *observations* $O$, and *actions* $A$.

By definition, "partially observable" means we cannot directly observe the state, instead we form a *belief state*, which is a probability distribution over the world states.

We must also define a *state transition probability distribution* $T(s, a, s')$, which is a conditional probability distribution describing probabilities of all outcome states $s' \in S$ given a starting state $s \in S$ and an action $a \in A$.

Next, we need an *observation probability distribution* $O(s, z)$ which describes the probabilities of perceiving observation $z \in O$ in state $s$. Note that unlike observation probability distributions of some POMDPs, ours are not also dependent on the most recent action.

By virtue of the *Markov property*, the transition and observation probability distribution functions are conditionally independent of all previous time-steps given the current time-step. The upshot is that our functions are time-independent, which greatly simplifies the model and our calculations.

Lastly, we must define a *utility function* $U(s): S \to \Re$, which assigns a *utility value* $r \in \Re$ to every state (Pineau et al. 2006). This will allow us to derive an *expected utility function* from the belief state to a reward value.

In summary, in order to implement an agent in a POMDP, we must start with a *prior estimation* (guess) of the initial belief state $p(s)$. Every time we make an observation, we can use the new information to *update* our belief state based on $O(s, z)$ with Bayes' rule to obtain the *posterior* belief state, $p(s|z)$. As an autonomous *agent*, we have the means to affect our state at will by taking one of multiple legal *actions*. We *predict* a new world state after each of the possible actions we can take using $T(s, a, s')$ and Bayes' rule. We choose to perform the action that maximizes our *expected utility*.

## TURKONTROL's POMDP Environment

We call the actual piece of work produced by a crowdworker the *artifact*. In TURKONTROL's formulation, the belief state is the joint probability distribution $f_{Q,Q'}(q,q')$ of the qualities of the two current artifacts; the artifact posted on the Web to be improved, and the one submitted as "improved" by a crowdworker. Each task posted on AMT for workers to complete is seen as an action taken by the agent. The three possible actions are:

1) **request improvement job** — hire a worker to improve upon the current state of the artifact
2) **request ballot job** — hire a worker to vote whether the new version of the artifact is in fact better than the previous version
3) **submit result** — stop iterating and treat the task as finished

As required in a POMDP, Dai et al. model the effects of these actions and observations on the belief state, then they define a utility function from quality to money. Now they can optimize the use of actions to get the best quality artifact for their money.

## TURKONTROL, More Concretely

An assortment of distributions and functions are used to model the effects of TURKONTROL's actions and update the functions themselves. Here, we shall review a few of the elements of TURKONTROL most important for the discussion of TURKPF.

One key assumption is the prior probability density function (PDF) for the quality $q$ of the current artifact $\alpha$, denoted $f_Q(q)$, and the prior PDF for the quality $q'$ of the "improved" artifact $\alpha'$, denoted $f_{Q'}(q')$. In this context, "an artifact with quality $q$ means an average dedicated worker has probability $1 - q$ of improving it" (DTC, pg. 1169), so its range is a real-number $q \in [0,1]$. TURKONTROL assumes the quality of the initial artifact is Beta distributed, because that fit the training data they used (Dai et al. 2011), and because it is the conjugate prior for Bernoulli random trials.

### Improvement Jobs

An update function is necessary to predict how $f_Q(q)$ becomes $f_{Q'}(q')$ after an improvement task is completed by worker $x$. In DTC, this is accomplished by computing the convolution

$$f_{Q'}(q') = \int_0^1 f_{Q'|q,x}(q') f_Q(q) dq \quad (1)$$

where $f_{Q'|q,x}(q')$ is a 2-dimensional distribution describing the conditional probability

$$f_{Q'|q,x}(q') = \mathbb{P}(Q' = q', Q = q \mid X = x) \quad (2)$$

In words, given that worker $x$ is the one performing the improvement task, we obtain a PDF of the artifact's new quality as a function of the original quality of the artifact.

**Ballot Jobs**

After each improvement job, we have two versions of the artifact, $\alpha$ and $\alpha'$. From (eq. 1) we have PDFs representing our understanding of the qualities of the artifacts $f_Q(q)$ and $f_{Q'}(q')$. Now we either want to request another improvement or submit the result. For either option, we would like to use the better of the two versions, but perhaps we are not confident that we know which version is better. This suggests we should ask a crowdworker for an evaluation, i.e. request a ballot job.

The specific formulation can be found in DTC, but note that the *utility of a ballot job* peaks when we are least sure which of $\alpha, \alpha'$ is superior.

## TURKONTROL $\Rightarrow$ TURKPF

We used a particle filter to model the probability distributions of the quality values of the newest version of an item $f_{Q'}(q')$, and the previous version $f_Q(q)$. The sets of particles in these particle filters will be denoted $\mathcal{P}'$ and $\mathcal{P}$ respectively.

Instead of maintaining objects representing the entire real-valued PDF $f_Q(q)$, we instantiate the prior Beta(1,9) distribution, and then sample from it $N$ times to obtain the initial state of our particle filter. The set of particles $\mathcal{P}$ is stored as an array of $\{ q_i \in [0,1] \}$, where each member of the array represents a different estimate of the true quality of the artifact.

**Some Assumptions**

First we assume that we have equal confidence in the ability of each particle to represent the truth, and that we have complete confidence that the particle filter does represent the truth. This allows us to assume

$$\mathbb{P}(Q = q_i) = \frac{1}{|\mathcal{P}|} \quad \forall q_i \in \mathcal{P} \qquad (3)$$

where $|\mathcal{P}| = N = $ *number of particles*.

Now we modify the equations from DTC to account for the fact that we cannot perform a normal integration over our particles, which do not represent a *function* over continuous range, but rather a *set* of values drawn from a continuous range. We replace DTC's integrals with a linear combination of the appropriate functions applied to each of the particles. All of the particles get a weight according to our confidence they are correct (eq. 4). For example, we can find the *expected utility of the unimproved artifact* via

$$\mathbb{E}\left[U\left(f_Q(q)\right)\right] = \sum_{q_i \in \mathcal{P}} \frac{U(q_i)}{|\mathcal{P}|} \qquad (4)$$

where we are summing up the utility values corresponding to all of the particles, and dividing by the number of particles.

**Improvement Job (Update Step)**

In order to compute (eq. 1) for the particle filter, we replace the convolution over $q$ with sampling:

---

*Particles* = $[q_1, \ldots, q_n]$; *NewParticles* = [ ]
**For** $q_i$ **in** *Particles*:
 $Dist = \text{Beta}\left(10\mu(q_i, \gamma), 10(1 - \mu(q_i, \gamma))\right)$
 *NewParticles*.append(*Dist*.sample())

---

As noted above, TURKONTROL's (eq. 1) updates a PDF $f_Q(q)$ by convolving it with a 2-dimensional distribution $f_{Q'|q,x}(q')$ (eq. 2) to obtain a new (1-dimensional) PDF $f_{Q'}(q')$. However in the particle filter algorithm, updating the distribution involves moving the particles to a new location based on their original location. Therefore, in the above algorithm, we apply a stochastic transformation to each particle, where the new location is a sample from a Beta distribution parameterized by the particle's initial value. Collectively, the resulting set of particles represents our estimate of the improved artifact.

This sampling method of propagation accomplishes the same thing as the original convolution. Eq. 1 says for every possible value of $q$, take the cross section of $f_{Q'|q,x}(q')$ (a Beta distribution), weight that cross section by the intensity of belief that this $q$ is the correct one, and sum those cross-sections up for all $q \in [0,1]$.

Now for the particle filter, $\forall q_i \in \mathcal{P}$ we find the cross-section of $f_{Q'|q,x}(q')$ (a Beta distribution). And (by eq. 3) we know we want each of these Beta distributions to have an equal impact on our resulting distribution of particles. As stated in the introduction, with infinite particles at any given $q$ we would end up with the exact distribution, which we can think of as what the original convolution is doing. But since randomly sampling from this Beta distribution gives us an unbiased approximation of its shape, we can approximate having infinite samples with a small number of particles. We assumed (eq. 3) that the influence of any particular particle should be the same, so each new particle is simply a sample from the distribution generated from the original value of an old particle. A formal proof of this equivalence would be more convincing and is probably straightforward, but this is the intuition.

**Ballot Job (Observation Step)**

In order to decide if a ballot job is the best use of money, we must find its expected utility, which means we must obtain posterior estimates of the artifacts after the vote, denoted $f_{Q|\overrightarrow{\mathbf{b}^{n+1}}}(q)$ and $f_{Q'|\overrightarrow{\mathbf{b}^{n+1}}}(q')$ (eq. 5, Appendix).

In the spirit of the basic particle filter algorithm, the mechanics of this convolution are different from the way we adapted the convolution in (eq. 1). Instead of taking each particle and moving it, we find the likelihood that each particle is correct according to the crowdworker's vote (the observation), $P(b_{n+1}|q')$ (eq. 7, Appendix). This will be the assigned *weight* of that particle. Now, a new set of $N$-particles is *resampled with replacement* from the original population, with each particle's selection-probability equal to its weight.

For example, say a crowdworker voted that $\alpha' > \alpha$, and our prior estimate of the quality after any votes we have already received, $f_{Q|\overrightarrow{\mathbf{b}^n}}(q)$, says that we already thought that $\alpha$ was of high quality. Now, any $q'_i \in \mathcal{P}'$ with a very low value is going to be assigned a very low weight because it is unlikely given our vote/observation. After this process, we can expect that the particles retained are the ones that more accurately represent the underlying truth.

On the first vote, the vote vector $\overrightarrow{\mathbf{b}^n}$ will be empty, leaving us $f_{Q'|\overrightarrow{\mathbf{b}^n}}(q') = f_{Q'}(q') = \mathcal{P}'$. Then $P(b_{n+1}|q, q')$ (eq. 8, Appendix) is a function taking as input one particle from each particle filter and returning the probability of a "yes" vote. To obtain the posterior $\mathcal{P}'$ after a vote, we get a vector of mappings from particle to weight, i.e. $\{q'_i \to P(b_{n+1}|q'_i)\} \; \forall q'_i \in \mathcal{P}'$. To obtain this vector for a "yes" vote, we perform the following pseudocode:

```
Weights = []
Particles = [q₁, …, qₙ]; ParticlesPrime = [q'₁, …, q'ₙ]
For q'ᵢ in ParticlesPrime:
    sum = 0
    For qⱼ in Particles:
        sum += P(b_{n+1}|qⱼ, q'ᵢ)
    Weights.append(sum)
Weights.normalize()
```

where `normalize` ensures that `weights.sum = 1`. Now we sample $N$ times from $\mathcal{P}'$ with $P(q'_i) :=$ `weights[i]` to obtain our new $\mathcal{P}'$.

## Experiments

We were unable to obtain the dataset the authors of DTC used to learn model parameters and equations, but we did still run experiments using a variety of default-settings to verify that our implementation matches TURKONTROL, as well as outcomes one would expect.

### Simulating Improvements

To simulate workers in these experiments we maintained "ground-truth" representations of the current states of the artifacts $\alpha, \alpha'$. During an improvement job, we used the `predict` function to assign a value for $\alpha'$ based on $\alpha$. To simulate a worker improving the artifact *believed* to be of higher-quality, we did the following:

```
If (meanParticleValue(f_Q(q)) <
    meanParticleValue(f_{Q'}(q')))
    α = α'
μ = find_improvementFunctionMean(α)
// average the result of three samples from Beta dist
α' = Beta(10·μ, 10·(1 − μ)).sample(3).sum / 3
```

### Number of Particles

The number of particles in each particle filter seems to have hardly if any effect on the final net utility (Figure 2). As $N$ increases exponentially from 6 to 7,180, our trendline for net utility increases (~linearly) from 400 to 480, but the range of variation is much wider than that. This lack of correlation is likely because although less particles means TURKPF makes more crude approximations, these approximations are still unbiased.

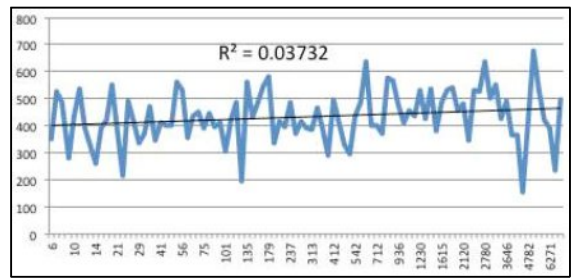

**Figure 2:** *Estimated Net Utility vs. Number of Particles. Trendline is Linear.*

## True Worker-Error

Similar to Dai et al's results, we see a decline in net utility as the simulated "true" $\gamma_{\bar{x}}$ increases (without changing the model's initial guess at $\gamma_{\bar{x}}$, Figure 3). The decline is much less drastic in our results than in those of DTC.

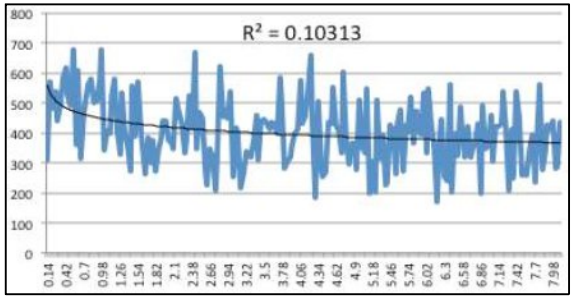

**Figure 3:** *Estimated Net Utility vs. $\gamma_{\bar{x}}$*

## Budget Size

There is a relationship between budget and final net utility only until the budget gets big enough that the algorithm submits the artifact before running out of money.

## Improvement Cost

As the cost of an improvement job increases relative to the cost of a ballot job, the number of improvement jobs falls (Figure 4). This was run using a budget of $100 and a ballot-job cost of $0.20 so that the money would not run out too quickly to get a good idea of the change occurring. The average final net utility remains constant as the improvement cost sweeps. The number of ballot jobs is too small to find a pattern in them.

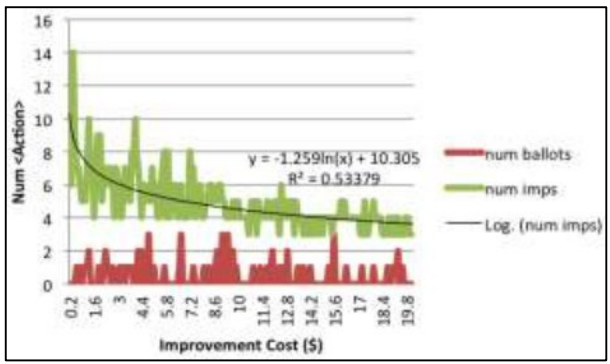

**Figure 4:** *Number of each action taken as improvement cost increases relative to ballot cost*

## Lookahead

Although using the Lookahead did produce different actions than not using it, we found no concrete relationship between Lookahead depth and the final net utility. This contrasts with the results in DTC, who found that for larger budgets, deeper Lookaheads lead to better performance. We don't know the source of this discrepancy.

## *"Dont_submit()"*

To understand what is happening during a typical run, we also ran the algorithm without letting it submit until running out of money, and Figures 5 and 6 were generated from this run. Seeing as the particle filter is a form of Monte Carlo simulation, the outcome is non-deterministic, and every run is slightly varied.

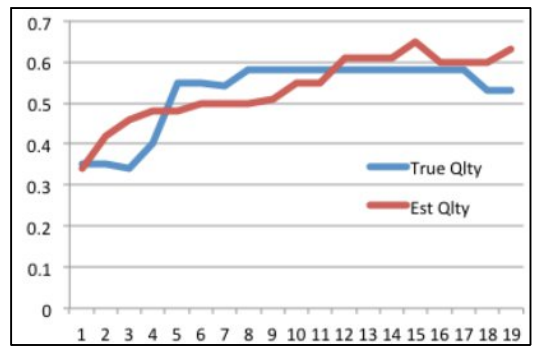

**Figure 5:** $\max\{\alpha, \alpha'\}$ vs. $\max\{\overline{f_Q(q)}, \overline{f_{Q'}(q')}\}$, where $\overline{f_Q(q)} = \frac{\Sigma_{q_i \in \mathcal{P}} q_i}{|\mathcal{P}|}$

| Action | True Utility | True Qlty | Est Qlty | Est-Real | Would've Submitted? |
|---|---|---|---|---|---|
| Imprvmt | 240.74 | 0.35 | 0.34 | -0.01 | |
| Imprvmt | 199.84 | 0.35 | 0.42 | 0.07 | |
| Imprvmt | 238.5 | 0.34 | 0.46 | 0.12 | |
| Imprvmt | 288.97 | 0.4 | 0.48 | 0.08 | |
| Imprvmt | 431.24 | 0.55 | 0.48 | -0.07 | |
| Imprvmt | 259.82 | 0.55 | 0.5 | -0.05 | |
| Imprvmt | 417.9 | 0.54 | 0.5 | -0.04 | |
| Imprvmt | 460.82 | 0.58 | 0.5 | -0.08 | |
| Imprvmt | 389.26 | 0.58 | 0.51 | -0.07 | |
| Ballot | 460.82 | 0.58 | 0.55 | -0.03 | |
| Imprvmt | 460.82 | 0.58 | 0.55 | -0.03 | yes |
| Ballot | 460.82 | 0.58 | 0.61 | 0.03 | |
| Imprvmt | 460.82 | 0.58 | 0.61 | 0.03 | yes |
| Imprvmt | 460.82 | 0.58 | 0.61 | 0.03 | yes |
| Ballot | 460.82 | 0.58 | 0.65 | 0.07 | |
| Ballot | 460.82 | 0.58 | 0.6 | 0.02 | yes |
| Ballot | 409.02 | 0.58 | 0.6 | 0.02 | |
| Imprvmt | 409.02 | 0.53 | 0.6 | 0.07 | yes |
| Ballot | 409.02 | 0.53 | 0.63 | 0.1 | yes |
| **Submit** | **409.02** | | | | |

**Figure 6:** *What happens when TURKPF cannot submit until it runs out of money*

# Conclusion/Future Work

Since we have not yet used Machine Learning on real data to determine appropriate values and functions for all of the default parameters and equations, TURKPF is not currently fit for use in a real world situation. Methods for obtaining these values and functions were described in (Dai et al. 2013), and follow well-known approaches for model-learning, such as Expectation Maximization. Aside from this and connecting to the AMT API, TURKPF is a functioning autonomous agent for allocating tasks to crowdworkers in an iterative improvement workflow. Iterative improvement is very flexible, for example it can be used in a Map-Reduce job-flow in either a Map or a Reduce task.

The reason DTC mentions the particle filter is for its speed advantages. Being able to choose the number of particles to use is a way to name your desired speed. With the model specifics we used, as described in the tables, an action is always chosen in much less than fone second, which is insignificant compared to the time it takes for a crowdworker to come across, accept, and complete a task. If one found it necessary to use large particle sets or deep Lookaheads, one could speed this up by using uniform discretization *and* a particle filter. If a complex update function is necessary, its execution is independent across particles, so it could be parallelized across multiple cores with the *parallel collections* in the Scala standard library.

In follow-on work by some of the same authors as DTC, AGENTHUNT improves on TURKONTROL by allowing the task designer to permit the agent to dynamically change its workflow (Lin et al. 2012). The example given has multiple possible observation-tasks: using an absolute rating scale, or changing the wording in the task description. The optimal observation task may vary over the course of improving the artifact, so AGENTHUNT is outfitted to learn how to choose the proper workflow offline and online. In experiments on AMT, this allows them to produce a better result with less money.

There are a number of further improvements that would make for interesting future work.

One might try making the average quality of both items

$$\frac{\sum_{p_i \in \{\mathcal{P} \cup \mathcal{P}'\}} p_i}{|\{\mathcal{P} \cup \mathcal{P}'\}|}$$

go down if the ballot is false, and up if the ballot is true. As implemented, the operation applied to update $\mathcal{P}$ and $\mathcal{P}'$ is completely symmetrical, so there is no change in the average quality of the two (aside from chance sampling variation). This adjustment would represent the fact that if the improved item is higher quality than the original, it seems more likely that improvements remain to be made than the update function predicted. As currently implemented, a sequence of true votes will generally lead TURKPF to believe that the improved item is of very high quality, and should be submitted because further improvement would be unlikely, wasting the opportunity.

In this vein, it would also be reasonable to have ballot jobs affect the `predict` function itself. As implemented, TURKPF's `predict` function has no "learning parameter" that gets updated online the way that the observe function has the worker error $\gamma$.

Also, we used the Beta distribution for the `predict` function (both for the model's estimation and the simulated crowdworkers) because that was function used in DTC. However, note that this implementation may not fit the scenario. For example, if the "ground-truth" quality $\alpha = 0.5$, then $\mu = 0.6$, meaning that $\mathbb{E}[\alpha'] = 0.6$, which contradicts the original definition of the quality of an item, which is $1 - \mathbb{P}(\textit{improvement by worker } \bar{x})$.

Of course, not obeying this definition is not problematic if it works in practice. However, this `predict` function may *not* correspond well to reality, because it has almost as much mass at reduced quality values as at improved quality values. This is counterintuitive because if a person was unable to improve an artifact, he or she would likely submit the artifact untouched or only slightly modified. In future work, it would be interesting to try a model that accounts for the assumption that a worker who is unlikely to improve an artifact is still unlikely to submit an artifact of *lesser* quality.

Finally, one could imagine a situation in which the user of TURKPF would want to prioritize getting very high quality artifacts, better than the average worker can produce. This would require a new `improvement_utility` function, because as-written it does not encourage improving artifacts that are unlikely to be improved.

**Acknowledgements.** We thank Aashish Sheshadri, Mausam, Peng Dai, Chris Lin, and the anonymous reviewers for their valuable feedback. This material is based upon work supported by the National Science Foundation, under Grant No. 1253413, and by a Temple Fellowship.

# Appendix

This is a quick reference to some of the equations used in the paper, these equations all originate from (Dai et al. 2010).

$$f_{Q'|\overline{\mathbf{b}^{n+1}}}(q') = \beta \cdot P(b_{n+1}|q') f_{Q'|\overline{\mathbf{b}^n}}(q') \quad (5)$$

where the $\beta$ implies that we must normalize $f_{Q'|\overline{\mathbf{b}^{n+1}}}(q')$ so that

$$\sum_{b_{n+1} \in \{True, False\}} \beta \cdot P(b_{n+1}|q') f_{Q'|\overline{\mathbf{b}^n}}(q') = 1 \quad (6)$$

$$P(b_{n+1}|q') = \int_0^1 P(b_{n+1}|q,q') f_{Q|\overline{\mathbf{b}^n}}(q) \, dq \quad (7)$$

$$P(b_{n+1}|q,q') = \begin{cases} a\big(d(q,q')\big) & \text{if } q' > q \\ 1 - a\big(d(q,q')\big) & \text{if } q' \leq q \end{cases} \quad (8)$$

$$a\big(d(q'q')\big) = \frac{1}{2}[1 + (1-d)^\gamma] \text{ for } \gamma > 0 \quad (9)$$

$$d(q,q') = 1 - |q - q'|^{\mathcal{M}} \quad (10)$$

where $\mathcal{M}$ is a constant set to $0.5$.